\documentclass[12pt,authoryear]{elsarticle} 
\usepackage{amsmath,amsfonts,amssymb}
\usepackage{algorithmic}
\usepackage{algorithm}
\usepackage{array}
\usepackage{subcaption}
\usepackage{textcomp}
\usepackage{stfloats}
\usepackage{url}
\usepackage{verbatim}
\usepackage{graphicx}
\usepackage{color}
\usepackage{tikz}
\usepackage{pgfplots}
\pgfplotsset{compat=1.18}
\usepackage{tikzviolinplots}
\usepackage{makecell}
\usepackage{scalerel}
\usepackage{orcidlink}

\usetikzlibrary{arrows.meta,arrows}
\usetikzlibrary{decorations.pathreplacing}
\usetikzlibrary{shapes.geometric}
\usetikzlibrary{svg.path}

\definecolor{MyRed}{HTML}{F94144}
\definecolor{MyDarkorange}{HTML}{F3722C}
\definecolor{MyOrange}{HTML}{F8961E}
\definecolor{MyYellow}{HTML}{F9C74F}
\definecolor{MyGreen}{HTML}{90BE6D}
\definecolor{MyTurquoise}{HTML}{43AA8B}
\definecolor{MyBlue}{HTML}{577590}

\definecolor{terrain00}{HTML}{333399}
\definecolor{terrain15}{HTML}{0099FF}
\definecolor{terrain25}{HTML}{00CC66}
\definecolor{terrain50}{HTML}{FFFF99}
\definecolor{terrain75}{HTML}{7F5C54}
\definecolor{terrain100}{HTML}{FFFFFF}

\hypersetup{urlcolor=cyan, colorlinks=true, linkcolor=blue}
\journal{ISPRS Journal of Photogrammetry and Remote Sensing}

\begin{document}

\begin{frontmatter}



\title{The Moon's Many Faces: A Single Unified Transformer for Multimodal Lunar Reconstruction} 


\author[1]{Tom Sander \orcidlink{0009-0008-3051-5976}\corref{cor1}}
\ead{tom.sander@tu-dortmund.de}

\author[1]{Moritz Tenthoff \orcidlink{0000-0003-2802-9561}}
\ead{moritz.tenthoff@tu-dortmund.de}

\author[1]{Kay Wohlfarth \orcidlink{0000-0002-4324-1459}}
\ead{kay.wohlfarth@tu-dortmund.de}

\author[1]{Christian Wöhler}
\ead{christian.woehler@tu-dortmund.de}

\cortext[cor1]{Corresponding author}

\affiliation[1]{organization={Image Analysis Group,
        TU Dortmund University},
        addressline={Otto-Hahn-Str. 6},
        city={Dortmund},
        postcode={44227},
        state={North Rhine-Westphalia},
        country={Germany}
}

\begin{abstract}
        Multimodal learning is an emerging research topic across multiple disciplines but has rarely been applied to planetary science. In this contribution, we propose a single, unified transformer architecture trained to learn shared representations between multiple sources like grayscale images, Digital Elevation Models (DEMs), surface normals, and albedo maps. The architecture supports flexible translation from any input modality to any target modality. Our results demonstrate that our foundation model learns physically plausible relations across these four modalities. We further identify that image-based 3D reconstruction and albedo estimation (Shape and Albedo from Shading) of lunar images can be formulated as a multimodal learning problem. Our results demonstrate the potential of multimodal learning to solve Shape and Albedo from Shading and provide a new approach for large-scale planetary 3D reconstruction. Adding more input modalities in the future will further improve the results and enable tasks such as photometric normalization and co-registration.
\end{abstract}




\end{frontmatter}

\section{Introduction}
Multimodal learning integrates information from heterogeneous data sources with distinct structures and statistical properties, such as images, spectral data, and text, to enhance performance on complex tasks. By leveraging complementary information across datasets, it can improve generalization and robustness \citep{4M2024,4M-21,Multi-MAE}. It has been widely applied in vision–language models \citep{dall-e,gpt,gpt-4}, healthcare \citep{multimodal-healthcare}, autonomous systems \citep{multimodal-autonomous}, and Earth remote sensing \citep{TerraMind,MMmesh}. In Earth observation, multimodal approaches have enabled sensor fusion \citep{sensor-fusion-overview}, denoising \citep{hyperspectral-denoising-cnn,hyperspectral-denoising-cnn2,denoising-remote-sensing}, gap filling \citep{gap-filling-unet,gap-filling-clouds}, cross-modal prediction \citep{remote-clip,geo-clip}, and geospatial visual question answering (GeoGPT \citep{geogpt}). Very recently, TerraMesh \citep{TerraMesh} demonstrated the application of multimodal learning and foundation models to large-scale Earth datasets.

Planetary science provides extensive heterogeneous datasets, but multimodal learning has been applied only rarely in this domain. Planetary datasets present unique challenges, including sparse spatial coverage, variable illumination, and limited ground truth. Multimodal and foundation model approaches could improve generalization, capture inter-dataset relationships, and support tasks such as generating planetary data products. These considerations motivate a systematic exploration of planetary remote sensing tasks that could benefit from multimodal learning.

A central challenge in planetary remote sensing is the joint estimation of Digital Elevation Models (DEMs) and albedo maps from reflectance or grayscale images, which is inherently ill-posed and computationally demanding. Traditional Shape and Albedo from Shading methods can address this problem but require substantial expertise and processing time. Reformulating it as a multimodal learning task enables simultaneous prediction of DEMs, albedo maps, and surface normals, producing more robust and computationally efficient results. This framework can be extended to larger-scale datasets, potentially supporting high-resolution regional or global DEM generation.

We introduce a transformer-based architecture that jointly learns shared representations across these modalities, explicitly modeling physically consistent relationships between surface geometry and reflectance. Incorporating additional inputs, such as multiple images or illumination geometry, can further enable photometric normalization, co-registration, and multi-mission data fusion, demonstrating the flexibility and extensibility of the proposed approach.

\section{Related works}
This section reviews previous work on reflectance models, which underpin common lunar remote sensing tasks such as reflectance normalization and photometric parameter retrieval. It then examines surface reconstruction approaches, emphasizing recent machine learning developments. Finally, we discuss the shift from unimodal representation learning to multimodal approaches and their applications in remote sensing.

\subsection{Reflectance Models}
A reflectance model is required to normalize measured reflectance data and retrieve surface properties.
Reflectance models can be divided into empirical functions and physically motivated analytical models \citep{shepard-2017}. 
The simplest approximation is the Lambert model. However, the lunar surface is not isotropic, as the reflectance depends strongly on the phase angle. 
Instead, the Lambert reflectance is often combined with a phase function and the Minnaert function \citep{minneart-1941} or Lommel-Seeliger law.
\cite{mcewen-1991} propose the empirical Lunar-Lambert model for photoclinometry and reflectance normalization. 
Analytical functions model the radiative light transfer and are more accurate but also more computationally expensive \citep{shepard-2017}.
The most prominent models are the \cite{skhuratov-1999} and \cite{hapke-2012} model.
Both model the scattering of light in a particulate medium, i.e., the lunar regolith.
The Hapke model is more widely used in planetary science and can therefore be assumed to be the standard model for lunar reflectance \citep{hess-2022}.
\cite{shepard-2007} perform extensive tests of the Hapke model and discuss the model inversion for parameter retrieval \citep{helfenstein-2011-inversion}.

\subsection{Surface Reconstruction}
Digital Elevation Model (DEM, sometimes also DTM) generation methods can be grouped into three broad categories: photogrammetry, ranging techniques, and intensity-based methods. More recently, machine learning approaches have emerged that use data from the other methods to generate new DEMs.

Photogrammetry or Stereo vision uses image pairs acquired to compute the disparity between tie-points found by block-matching \citep{beyer-2018-asp}.
The disparities and known camera parameters are used to find the 3D coordinates of the tie-points and interpolate a DEM.
Stereo vision is well suited for the creation of global DEMs with high vertical accuracy. 
However, it is limited by the availability of suitable stereo pairs and the matching process.
Due to the size of the matching window, the Ground Sampling Distance (GSD) of the resulting Stereo DEM is typically chosen 3-5 times larger than the GSD of the input images \citep{beyer-2018-asp,kirk-2021-comparison}.
The actual resolution of the DEM is even coarser due to matching and interpolation errors \citep{kirk-2017-isprs,beyer-2018-asp}.
Typical reconstruction artefacts include holes in shadowed areas and stair-like structures due to pixel locking \citep{gehrig-2016}.
The most commonly used frameworks for Stereo DEM generation are the open source Ames Stereo Pipeline (ASP) \citep{beyer-2018-asp} and the proprietary SOCET SET from BAE Systems \citep{miller-1993-socet}. 
Ranging techniques include laser altimetry (e.g., the Lunar Orbiter Laser Altimeter (LOLA) \citep{smith-2010-lola}) and radar interferometry \citep{hensley-2008-interferometry}.
The instrument sends out a signal and measures the return time of the reflected signal to determine the distance to the surface. 
Due to the active nature of the method, it is not dependent on the illumination conditions and can also measure in permanently shadowed regions.
The sampling points provide very high vertical accuracy, but the resolution of derived DEMs is limited by the comparatively large cross-track distance between tracks \citep{barker-2016} and interpolation artefacts.
\cite{barker-2016} combined LOLA data with stereo DEMs from the SELENE Terrain Camera \citep{haruyama-2008-tc} to create the SLDEM512 (about $60\,\mathrm{m}/\mathrm{pixel}$). 
Intensity-based methods relate the intensity of a pixel to the surface orientation and the illumination conditions. 
Notable early contributions include the works of \cite{van-diggelen-1951}, \cite{rindfleisch-1966}, \cite{wildey-1975}, and \cite{kirk-1987}, who utilized profile- or area-based photoclinometry to determine the shape of planetary surfaces. 

Photoclinometry estimates the surface slope, which is subsequently integrated to obtain the surface height.
In contrast, Shape from Shading (SfS) methods estimate the surface slope and height simultaneously \citep{horn1990-sfs,dem-grumpe2014}.
However, the terms are often used interchangeably in the literature.
Because the reconstruction is based on the surface orientation, the resulting surface slope is very accurate. 
However, no absolute height information can be retrieved from a single image, and cumulative small integration errors can lead to large deviations from the true surface height \citep{dem-grumpe2014}.
To increase the vertical accuracy \cite{shao1991-sfs} introduced a depth constraint to tie the estimated DEM to a lower resolution initialization.
The fusion of SfS with Stereo DEMs and laser altimetry data has been continuously improved over the following years \citep{soderblom-kirk-2003,fassold-2004,lohse-2006} which the work by Grumpe and Wöhler \citep{dem-woehler2012,dem-grumpe2014,dem-grumpe2014-2} builds on by introducing a more generalized variational approach. They present a new framework that simultaneously estimates surface heights and local reflectance properties, such as the single-scattering albedo, using the Hapke reflectance model \citep{hapke-2012}. Instead of a smoothness constraint, that tends to oversmooth the surface, they use an absolute and relative depth constraint. We utilize this extensively validated \citep{dem-grumpe2014,tenthoff-2020,hess-2022} Shape and Albedo from Shading framework to generate DEMs and albedo maps for our multimodal model.
Other SfS frameworks \citep{alexandrov-2018-ames-sfs,wu-2018,jiang-2017, kirk-2022-compare} use similar approaches to compute DEMs, but employ either simpler reflectance models, smoothness constraints or a fixed albedo.
To the best of the authors' knowledge, only the framework introduced by \citet{dem-grumpe2014-2} and extended by \citet{tenthoff-2020} and \citet{hess-2022} is capable of jointly estimating the surface height and single-scattering albedo from a single image.

Machine learning techniques can directly infer a relationship between a measured image and a corresponding DEM, created by the methods described above.
The trained network can then be used to create entirely DEMs in much less time than traditional methods.
\cite{demCNN1,demCNN2} use a low-resolution DEM and a high-resolution grayscale image as input to learn a transformation that generates new high-resolution DEMs. To achieve this, the authors utilize two convolutional neural networks (CNN), each representing a different modality, and concatenate the features from these streams to a decoder that generates the DEM. The authors extended their model in \citep{chen-2024-transformer} and exchanged the high-resolution image branch with a Swin Transformer network \citep{swinTransformer}. In contrast, our approach unifies multiple paths from different sources into a single model. This means our transformer is not restricted to generating high-resolution DEMs solely from a given grayscale image and an initial low-resolution DEM. In fact, we do not require an initial surface to generate our DEMs. Our broader goal is to learn the correlations between different representations within a single model.

\cite{chen-2021-otherChen} developed Mars3DNet, a CNN-based model designed to generate DEMs of the Martian surface. The model processes the Context Camera (CTX) images as input and employs an auto-denoising network to create a Lambertian image. It then predicts a DEM from this Lambertian image and scales the prediction using an initial low-resolution DEM. In contrast, our approach utilizes a Transformer framework and an any-to-any methodology, allowing our network to produce a broader range of outputs beyond just DEMs.

Tao et al. utilize Generative Adversarial Network (GAN) models in order to create MADNet 1.0 \citep{tao-2021-madnet1} and MADNet 2.0 \citep{tao-2021-madnet2} to generate high-resolution DEMs of the Moon \citep{tao-2023-moon} and Mars \citep{tao-2023-mars}. In their approach, the authors employ a convolutional U-Net as the generator and a CNN as the discriminator to estimate relative heights from the provided input image. The absolute heights are then obtained by rescaling the predicted relative heights using a low-resolution initial DEM. While some methods obtain absolute heights by rescaling the predicted relative heights using a low-resolution initial DEM, other GAN-based approaches use the low-resolution DEM as a direct input to the network \citep{liu-2022-gansfs, chen-2025-lunarDEM}.

A further GAN approach proposed by \cite{cao-2024} and \cite{yang-2024-attention} combines the U-Net GAN with an attention mechanism utilizing the channel and spatial attention in the U-Net network. 

Our approach generally differs from previous models by employing a Transformer architecture, which allows for greater flexibility and enhanced correlation learning across multiple modalities. Unlike the CNN-based models, which are typically limited to generating specific outputs such as DEMs from single grayscale images, our model supports any-to-any generation synthesizing varying outputs, including Normal, Albedo, DEM, and more. 
This allows us to learn the capabilities of the SfS framework of \citet{dem-grumpe2014-2}.
Furthermore, our approach differs by explicitly separating albedo and height information, treating them as two distinct modalities. This holistic perspective broadens the scope of our work, allowing us to investigate and learn inter-modal relationships more effectively.

\subsection{Multimodal Representation Learning for Remote Sensing}
This paper aims to develop a unified representation from various lunar low-level processing tasks by building upon and extending existing representation learning methodologies. Representation learning techniques, such as autoencoders \citep{hinton_autoencoder}, CNNs \citep{lecun_nature}, and VAEs \citep{kingma_vae}, extract meaningful features from raw data by reconstructing inputs or learning data distributions, with newer methods like masked autoencoders \citep{MAE} excelling at inferring missing information. These approaches have been effectively applied to remote sensing and lunar science, enabling tasks like hyperspectral image classification \citep{hyperspectral_repr}, thermal pattern modeling \citep{ThermalVAE}, surface anomaly detection \citep{MoonAnomaly}, and crater analysis through transfer learning \citep{crater_detection, crater_detection_tu}.

The field of multimodal learning involves the extension of representation learning to manage multiple data types simultaneously. A common strategy is to convert different modalities into a shared representation format. This often involves the use of modality-specific tokenizers or projection layers, followed by processing through a unified model architecture. Approaches like Multi-MAE \citep{Multi-MAE} utilize shared transformers on projected embeddings from different modalities, while others like 4M \citep{4M2024, 4M-21} employ VQ-VAE \citep{VQVAE} tokenizers before utilizing a masked encoder-decoder setup (similar concepts underpin models like Unified-IO \citep{Unified-IO1, Unified-IO2} for broader data types).

Multi-modal learning is increasingly explored in Earth observation, with works like MMEarth \citep{MMmesh}, TerraMesh \citep{TerraMesh}, and TerraMind \citep{TerraMind} integrating diverse geospatial earth related datasets for enhanced understanding. However, applying such sophisticated multimodal fusion techniques specifically to lunar remote sensing data is a emerging area. Our previous work \citep{AGUAbstract, LPSCAbstract} represents some of the initial efforts in leveraging machine learning to uncover relationships between disparate lunar datasets, with a particular focus on mapping lunar hydration. This paper represents an important initial step in investigate multimodal analysis within lunar studies by developing a more comprehensive multimodal framework for unified low-level task representation.


\section{Dataset}
\begin{table}
	\centering
	\begin{tabular}{l c c c c}
		\hline
		\thead{Site \\ name:}	& Image ID: 	& \thead{Incidence \\ angle [\textdegree]:}	& \thead{Phase \\ angle [\textdegree]:}	& \thead{Nominal \\ Resolution [m]:} \\\hline\hline		
		Apollo 16	& M144524996	& $47.18$							& $50.98$						& $0.5$								 \\
		Apollo 17	& M168000580	& $45.13$							& $71.15$						& $0.5$								 \\
		Apollo 11	& M175124932	& $40.98$							& $39.86$						& $0.5$							     \\
		Apollo 15	& M175252641	& $49.39$							& $48.43$						& $0.5$							     \\
		Apollo 14	& M175388134	& $44.91$							& $29.33$						& $0.5$								 \\
		Apollo 12	& M175428601	& $45.16$							& $44.03$						& $0.5$								 \\\hline
	\end{tabular}
	\caption{All NAC images used in this work are constructed by stitching together both left and right side NAC images and normalizing them using ISIS3 \citep{isis3}.}
	\label{tab:data:NACImage}
\end{table}
Our dataset consists of six large narrow-angle camera (NAC) images from the Lunar Reconnaissance Orbiter (LRO) \citep{LRO1, LRO2}. Specifically, these NAC images correspond to the Apollo 11 through 17 landing sites. 
These site selections are strategic, as they are among the most studied locations on the Moon, providing ground-truth data for validating our model. The Apollo sites represent a diverse range of lunar geology, including the rugged Descartes Highlands of Apollo 16, the unique Fra Mauro formation from Apollo 14, and the volcanic features of Apollo 15 and 17. This variety ensures that our model is trained on a broad representation of the lunar surface, enhancing its generalizability. 
For the creation of our training dataset, we included images from Apollo 12, 14, 15, 16, and 17 as training samples for our tokenizers and model. By combining and slicing these large images into patches measuring $224\times 224$ pixels, with a stride of $32$, we generated a total of $1,720,540$ images. Since the tokenizers operate with a patch size of 8, this results in $784$ tokens per image. As a result, the multimodal model was trained on over $1.34$ billion tokens. For testing, we used the NAC image from the Apollo 11 landing site, which yielded $22,405$ images to evaluate the algorithm, resulting in over $17$ million tokens per modality to test with.

We used our Shape and Albedo from Shading (SfS) framework to derive DEMs, albedo maps, and normal maps from the NAC images.
The following paragraph will briefly explain the method which we previously validated for the Moon \citep{dem-grumpe2014,dem-grumpe2014-2}, Mercury \citep{tenthoff-2020}, and Mars \citep{hess-2022}. A more detailed description of the method can be found in \citep{dem-grumpe2014,dem-grumpe2014-2}.

SfS requires a reflectance model $R$ that relates the shading observed in the image $I$ to the surface shape, i.e. the DEM representing the surface. 
A DEM is a function $z(x,y)$ that assigns a height to each pixel $(x,y)$. 
The components of the surface gradient are the partial derivatives of $z$ with respect to $x$ and $y$, respectively:
\begin{equation}
    p = \frac{\partial z}{\partial x} \quad \textnormal{and} \quad q = \frac{\partial z}{\partial y}
\end{equation}
The surface normal vector $\mathbf{n}$ is defined as
\begin{equation}
    \mathbf{n} = \frac{1}{\sqrt{p^2+q^2+1}} 
    \begin{pmatrix} 
    -p \\ 
    -q \\ 
    1 
    \end{pmatrix}
\end{equation}
The surface slope angle $\alpha$ is useful for hazard assessment maps and can be derived from the surface gradient or normal vector with
\begin{equation}
    \alpha = \arctan(\sqrt{p^2+q^2}) = \arccos(n_z)   
\end{equation}

The illumination and viewing geometry is defined by the incidence angle $i$ (between $\mathbf{n}$ and $\mathbf{s}$ pointing to the Sun), the emission angle $e$ (between $\mathbf{n}$ and $\mathbf{v}$ pointing to the camera), and the phase angle $g$ (between $\mathbf{s}$ and $\mathbf{v}$). 
Instead of a simple empirical model (e.g., Lambert or Lunar-Lambert), we opt to use the more complex Hapke model \citep{hapke-2012}. 
The Hapke model is a physically based reflectance model that works well for particulate surfaces, such as the lunar regolith.
We use the Anisotropic Multiple Scattering Approximation (AMSA) \citep{hapke-2012} of the Hapke model given by 
\begin{equation}
    \begin{split}
    R_\textnormal{AMSA}(i, e, g, w) &= \frac{w}{4\pi} \frac{\mu_0}{\mu_0 + \mu} \left[ \right. p(g)B_\textnormal{SH}(g)\\ &\left. + M(\mu_0, \mu) \right]  B_\textnormal{CB}(g) S(i,e,g,\bar\theta)
    \end{split}
\end{equation}  
with $\mu_0 = \cos(i)$ and $\mu = \cos(e)$. 
The single-scattering albedo $w$ is the intrinsic reflectivity of the surface material independent of the orientation. Our SfS framework allows us to estimate the pixelwise albedo $w$ simultaneously with the DEM.
The resulting albedo map is used for the multimodal model.
We use fixed values estimated by \cite{warell-2004} for the remaining parameters of the model.
The phase function $p(g)$ is given by the Double Henyey-Greenstein function with parameters $b=0.21$ and $c=0.7$ \citep{warell-2004}. We use $B_\textnormal{S0}=3.1$ and $h_\textnormal{S}=0.11$ for the Shadow Hiding Opposition Effect $B_\textnormal{SH}$.

Every SfS algorithm is based on the minimization of the intensity error 
\begin{equation}
    E_\textnormal{I} = \frac{1}{2} \int\limits_{x}\int\limits_{y} \left(R(p,q) - I\right)^2 dx dy
\end{equation}
between the modeled reflectance image $R$ and the observed image $I$.

\begin{figure}
    \centering
    \input{figures/tikz/sfs/pipeline.tex}
    \caption{The SfS algorithm requires a low-resolution DEM and a grayscale image as inputs. The output high-resolution DEM, surface normals, and an albedo map are used as input for the foundation model.} 
    \label{fig:sfs-overview}
\end{figure}
SfS is an ill-posed problem, because two unknowns $p$ and $q$ have to be estimated from a single pixel.
Therefore, additional regularization terms are required to obtain a solution. 
The error terms for the SfS algorithm that was used in this work are described in \ref{appendix:SfS}. 
The minimization of the resulting total error is performed using an iterative relaxation approach. The update equations can be found in \citep{dem-grumpe2014}. 
Figure \ref{fig:sfs-overview} shows the input and output of the SfS algorithm.
We use the output albedo map, the DEM, and the derived surface normals training and test data for our foundation model.

\section{Unified Transformer Approach}
\label{sec:UnifiedTransformer}
In this section, we present a systematic and detailed description of the machine learning methods used. The section is structured into three parts. We start with an overview that explains the overall goal. Next, we delve into the tokenizer section, which outlines how the tokenizers were built and trained. Finally, we discuss the masked autoencoding section, which describes the training procedure of the masked autoencoder.

\subsection{General idea}
To effectively manage various modalities, such as different sensor inputs, data types, and data sizes within a single model, it is essential to unify these modalities into a consistent dimension. This can be achieved by tokenizing each modality using a modified VQ-Tokenizer \citep{VQVAE}. After tokenization, the model is trained in a masked manner to predict the remaining tokens based on a set of unmasked tokens. By training these tokens in a cross-modal fashion, the model learns the different relationships between all modalities. This approach enables the model to understand how to transform, for example, a grayscale image into a DEM.

\subsection{Tokenization}
\begin{figure}
	\centering
	\input{figures/tikz/tokenization/vq-tokenization.tex}
	\caption{Custom VQ-Tokenizer \citep{VQVAE}: The input image is decoded using a vision transformer \citep{ViT}. The encoded embeddings are then reshaped to form a cube. After vector quantization, the input cube is decoded using a convolutional neural network.}
	\label{fig:method:vq}
\end{figure}
We use a custom-built tokenizer structure for the tokenization process, as shown in Figure \ref{fig:method:vq}. Specifically, a vision transformer \citep{ViT} serves as the encoder, while a convolutional neural network (CNN) acts as the decoder to reconstruct the image. The vision transformer, functioning as the encoder, generates a set of embeddings that are reshaped into a three-dimensional cube. This cube is then discretized using a vector quantizer. The quantized vectors within the cube are subsequently processed through the convolutional neural network to reproduce the original input image.
We use a Vision Transformer (ViT) \citep{ViT} as the encoder in combination with the vector quantization (VQ) \citep{VQVAE} tokenizer because of its ability to extract rich and contextual features from images. By utilizing the self-attention mechanism, the ViT effectively captures long-range dependencies and non-local relationships among different image parts. This results in more informative and robust feature representations. Consequently, our VQ tokenizer can learn a more effective and compact representation of the input image, which is crucial for achieving high-quality image reconstruction and compression.

\subsection{Masked Autoencoding}
\label{sec:method:maskedAutoencoder}
\begin{figure}
	\centering
	\begin{tikzpicture}
    \node[align=center] at (-3.6,0.4) {\small Modality 1};
    \foreach \i in {1,...,4}{
        \draw[fill=MyBlue!55!white, draw=MyBlue!50!black] (-5.0+\i*0.3, 0.0) rectangle (-5.0+\i*0.3+0.25, 0.25);
    }
    \node[align=center] at (-3.2, 0.125) {$\cdots$};
    \draw[fill=MyBlue!55!white, draw=MyBlue!50!black] (-2.9, 0.0) rectangle (-2.65, 0.25);
    \draw[draw=MyBlue!50!black, thick] (-4.7, -0.1) -- (-2.65, -0.1);

    \node[align=center] at (-0.6,0.4) {\small Modality 2};
    \foreach \i in {1,...,4}{
        \draw[fill=MyGreen!55!white, draw=MyGreen!50!black] (-2.0+\i*0.3, 0.0) rectangle (-2.0+\i*0.3+0.25, 0.25);
    }
    \node[align=center] at (-0.2, 0.125) {$\cdots$};
    \draw[fill=MyGreen!55!white, draw=MyGreen!50!black] (0.1, 0.0) rectangle (0.35, 0.25);
    \draw[draw=MyGreen!50!black, thick] (-2.0+0.3, -0.1) -- (0.35, -0.1);

    \node[align=center] at (2.35, 0.4) {\small Modality 3};
    \foreach \i in {1,...,4}{
        \draw[fill=MyRed!55!white, draw=MyRed!50!black] (1.0+\i*0.3, 0.0) rectangle (1.0+\i*0.3+0.25, 0.25);
    }
    \node[align=center] at (2.8, 0.125) {$\cdots$};
    \draw[fill=MyRed!55!white, draw=MyRed!50!black] (3.1, 0.0) rectangle (3.35, 0.25);
    \draw[draw=MyRed!50!black, thick] (1.0+0.3, -0.1) -- (3.35, -0.1);

    \draw[draw=MyBlue!50!black, thick] (-3.6, -0.1) -- (-3.6, -0.5);
    \draw[draw=MyGreen!50!black, thick, -latex] (-0.6, -0.1) -- (-0.6, -1.0);
    \draw[draw=MyRed!50!black, thick] (2.45, -0.1) -- (2.45, -0.7);

    \draw[draw=MyBlue!50!black, thick] (-3.7, -0.5) -- (2.25, -0.5);
    \draw[draw=MyGreen!50!black, thick] (-3.6, -0.6) -- (2.35, -0.6);
    \draw[draw=MyRed!50!black, thick] (-3.5, -0.7) -- (2.45, -0.7);

    \draw[draw=MyBlue!50!black, thick, -latex] (-3.7, -0.5) -- (-3.7, -1.0);
    \draw[draw=MyGreen!50!black, thick, -latex] (-3.6, -0.6) -- (-3.6, -1.0);
    \draw[draw=MyRed!50!black, thick, -latex] (-3.5, -0.7) -- (-3.5, -1.0);

    \draw[draw=MyBlue!50!black, thick, -latex] (-0.7, -0.5) -- (-0.7, -1.0);
    \draw[draw=MyRed!50!black, thick, -latex] (-0.5, -0.7) -- (-0.5, -1.0);

    \draw[draw=MyBlue!50!black, thick, -latex] (2.25, -0.5) -- (2.25, -1.0);
    \draw[draw=MyGreen!50!black, thick, -latex] (2.35, -0.6) -- (2.35, -1.0);
    \draw[draw=MyRed!50!black, thick, -latex] (2.45, -0.7) -- (2.45, -1.0);

    \node[align=center] at (-3.6, -1.2) {\footnotesize $\sim\mathrm{Dir}(\alpha = 0.1)$};
    \node[align=center] at (-0.6, -1.2) {\footnotesize $\sim\mathrm{Dir}(\alpha = 1.0)$};
    \node[align=center] at (2.35, -1.2) {\footnotesize $\sim\mathrm{Dir}(\alpha \rightarrow \infty)$};
    
    \foreach \i in {1,...,9}{
        \draw[fill=MyBlue!55!white, draw=MyBlue!50!black] (-5.25+\i*0.3, -1.5) rectangle (-5.25+\i*0.3+0.25, -1.75);
        \draw[fill=MyRed!55!white, draw=MyRed!50!black] (-5.25+\i*0.3, -1.8) rectangle (-5.25+\i*0.3+0.25, -2.05);
        \draw[fill=MyRed!55!white, draw=MyRed!50!black] (-5.25+\i*0.3, -2.1) rectangle (-5.25+\i*0.3+0.25, -2.35);
        \draw[fill=MyBlue!55!white, draw=MyBlue!50!black] (-5.25+\i*0.3, -2.4) rectangle (-5.25+\i*0.3+0.25, -2.65);
        \draw[fill=MyGreen!55!white, draw=MyGreen!50!black] (-5.25+\i*0.3, -2.7) rectangle (-5.25+\i*0.3+0.25, -2.95);
        \draw[fill=MyGreen!55!white, draw=MyGreen!50!black] (-5.25+\i*0.3, -3.3) rectangle (-5.25+\i*0.3+0.25, -3.55);
        \draw[fill=MyBlue!55!white, draw=MyBlue!50!black] (-5.25+\i*0.3, -3.6) rectangle (-5.25+\i*0.3+0.25, -3.85);
    }
    \draw[fill=MyGreen!55!white, draw=MyGreen!50!black] (-4.95, -3.0) rectangle (-4.7, -3.25);
    \draw[fill=MyGreen!55!white, draw=MyGreen!50!black] (-4.65, -3.0) rectangle (-4.4, -3.25);
    \draw[fill=MyGreen!55!white, draw=MyGreen!50!black] (-4.35, -3.0) rectangle (-4.1, -3.25);
    \foreach \i in {1,...,6}{
        \draw[fill=MyRed!55!white, draw=MyRed!50!black] (-4.35+\i*0.3, -3.0) rectangle (-4.1+\i*0.3, -3.25);
    }

    \draw[fill=MyGreen!55!white, draw=MyGreen!50!black] (-1.95, -1.5) rectangle (-1.7, -1.75);
    \draw[fill=MyGreen!55!white, draw=MyGreen!50!black] (-1.65, -1.5) rectangle (-1.4, -1.75);
    \draw[fill=MyGreen!55!white, draw=MyGreen!50!black] (-1.35, -1.5) rectangle (-1.1, -1.75);
    \foreach \i in {4,...,9}{
        \draw[fill=MyRed!55!white, draw=MyRed!50!black] (-2.25+\i*0.3, -1.5) rectangle (-2.25+\i*0.3+0.25, -1.75);
    }
    \draw[fill=MyBlue!55!white, draw=MyBlue!50!black] (-1.95, -1.8) rectangle (-1.7, -2.05);
    \draw[fill=MyBlue!55!white, draw=MyBlue!50!black] (-1.65, -1.8) rectangle (-1.4, -2.05);
    \draw[fill=MyBlue!55!white, draw=MyBlue!50!black] (-1.35, -1.8) rectangle (-1.1, -2.05);
    \draw[fill=MyGreen!55!white, draw=MyGreen!50!black] (-1.05, -1.8) rectangle (-0.8, -2.05);
    \draw[fill=MyGreen!55!white, draw=MyGreen!50!black] (-0.75, -1.8) rectangle (-0.5, -2.05);
    \draw[fill=MyGreen!55!white, draw=MyGreen!50!black] (-0.45, -1.8) rectangle (-0.2, -2.05);
    \draw[fill=MyRed!55!white, draw=MyRed!50!black] (-0.15, -1.8) rectangle (0.1, -2.05);
    \draw[fill=MyRed!55!white, draw=MyRed!50!black] (0.15, -1.8) rectangle (0.4, -2.05);
    \draw[fill=MyRed!55!white, draw=MyRed!50!black] (0.45, -1.8) rectangle (0.7, -2.05);
    \draw[fill=MyBlue!55!white, draw=MyBlue!50!black] (-1.95, -2.1) rectangle (-1.7, -2.35);
    \draw[fill=MyBlue!55!white, draw=MyBlue!50!black] (-1.65, -2.1) rectangle (-1.4, -2.35);
    \draw[fill=MyGreen!55!white, draw=MyGreen!50!black] (-1.35, -2.1) rectangle (-1.1, -2.35);
    \foreach \i in {4,...,9}{
        \draw[fill=MyRed!55!white, draw=MyRed!50!black] (-2.25+\i*0.3, -2.1) rectangle (-2.25+\i*0.3+0.25, -2.35);
    }
    \draw[fill=MyBlue!55!white, draw=MyBlue!50!black] (-1.95, -2.4) rectangle (-1.7, -2.65);
    \draw[fill=MyBlue!55!white, draw=MyBlue!50!black] (-1.65, -2.4) rectangle (-1.4, -2.65);
    \draw[fill=MyBlue!55!white, draw=MyBlue!50!black] (-1.35, -2.4) rectangle (-1.1, -2.65);
    \draw[fill=MyGreen!55!white, draw=MyGreen!50!black] (-1.05, -2.4) rectangle (-0.8, -2.65);
    \draw[fill=MyGreen!55!white, draw=MyGreen!50!black] (-0.75, -2.4) rectangle (-0.5, -2.65);
    \draw[fill=MyGreen!55!white, draw=MyGreen!50!black] (-0.45, -2.4) rectangle (-0.2, -2.65);
    \draw[fill=MyGreen!55!white, draw=MyGreen!50!black] (-0.15, -2.4) rectangle (0.1, -2.65);
    \draw[fill=MyRed!55!white, draw=MyRed!50!black] (0.15, -2.4) rectangle (0.4, -2.65);
    \draw[fill=MyRed!55!white, draw=MyRed!50!black] (0.45, -2.4) rectangle (0.7, -2.65);
    \draw[fill=MyBlue!55!white, draw=MyBlue!50!black] (-1.95, -2.7) rectangle (-1.7, -2.95);
    \draw[fill=MyBlue!55!white, draw=MyBlue!50!black] (-1.65, -2.7) rectangle (-1.4, -2.95);
    \draw[fill=MyBlue!55!white, draw=MyBlue!50!black] (-1.35, -2.7) rectangle (-1.1, -2.95);
    \foreach \i in {4,...,9}{
        \draw[fill=MyRed!55!white, draw=MyRed!50!black] (-2.25+\i*0.3, -2.7) rectangle (-2.25+\i*0.3+0.25, -2.95);
    }
    \foreach \i in {1,...,9}{
        \draw[fill=MyBlue!55!white, draw=MyBlue!50!black] (-2.25+\i*0.3, -3.0) rectangle (-2.25+\i*0.3+0.25, -3.25);
    }
    \foreach \i in {1,...,5}{
        \draw[fill=MyBlue!55!white, draw=MyBlue!50!black] (-2.25+\i*0.3, -3.3) rectangle (-2.25+\i*0.3+0.25, -3.55);
    }
    \foreach \i in {5,...,9}{
        \draw[fill=MyRed!55!white, draw=MyRed!50!black] (-2.25+\i*0.3, -3.3) rectangle (-2.25+\i*0.3+0.25, -3.55);
    }
    \foreach \i in {1,...,4}{
        \draw[fill=MyBlue!55!white, draw=MyBlue!50!black] (-2.25+\i*0.3, -3.6) rectangle (-2.25+\i*0.3+0.25, -3.85);
    }
    \foreach \i in {4,...,9}{
        \draw[fill=MyGreen!55!white, draw=MyGreen!50!black] (-2.25+\i*0.3, -3.6) rectangle (-2.25+\i*0.3+0.25, -3.85);
    }

    \foreach \j in {0,...,7}{
        \foreach \i in {1,...,3}{
            \draw[fill=MyBlue!55!white, draw=MyBlue!50!black] (0.75+\i*0.3, -1.5-\j*0.3) rectangle (0.75+\i*0.3+0.25, -1.75-\j*0.3);
        }
        \foreach \i in {4,...,6}{
            \draw[fill=MyGreen!55!white, draw=MyGreen!50!black] (0.75+\i*0.3, -1.5-\j*0.3) rectangle (0.75+\i*0.3+0.25, -1.75-\j*0.3);
        }
        \foreach \i in {7,...,9}{
            \draw[fill=MyRed!55!white, draw=MyRed!50!black] (0.75+\i*0.3, -1.5-\j*0.3) rectangle (0.75+\i*0.3+0.25, -1.75-\j*0.3);
        }
    }

    \node[align=center, rotate=90] at (-3.6, -4.1) {$\cdots$};
    \node[align=center, rotate=90] at (-0.6, -4.1) {$\cdots$};
    \node[align=center, rotate=90] at (2.35, -4.1) {$\cdots$};

\end{tikzpicture}
	\caption{Figure illustrates the Dirichlet sampling process for a set of three distinct modalities, similar to the approach described in \citep{Multi-MAE}. When the $\alpha$ parameter is low, only one modality, or occuasionally just a few tokens from a single modality, is selected for training, while the remaining tokens serve as targets. When $\alpha$ is set to $1.0$, a diverse mixture of tokens from all modalities is sampled each time. Conversely, as $\alpha$ approaches $\infty$, the sampling results in a uniform distribution across all modalities.}
	\label{fig:method:dirichletSampling}
\end{figure}
Developing an effective masking strategy to train and obtain a robust and meaningful representation of input data is essential. For this purpose, we employ the same masking strategy as detailed in \citep{Multi-MAE,4M2024,4M-21}. This strategy, illustrated in Figure \ref{fig:method:dirichletSampling}, involves a Dirichlet sampling process with a hyperparameter $\alpha$. When $\alpha$ is set to a small value, the process primarily samples from a single modality and aims to predict the remaining modalities. If $\alpha$ is close to 1, the sampled input token sequences will be more diverse. Conversely, as $\alpha$ increases, the resulting input token sequences converge towards a uniform sampling of all available modalities.

Another important aspect of the multi-modal masked autoencoder is pre-training. This step involves training one modality using another, which helps in learning a representation of these two initial modalities. The pre-training is conducted on a smaller model, while the other modalities are fine-tuned into a slightly larger model with an increased sequence length after the pre-training is completed.

\begin{figure*}
	\centering
	\input{figures/tikz/maskedAutoencoder/mae.tex}
	\caption{The figure illustrates the training process of the Multimodal Masked Autoencoder. Initially, different input modalities are tokenized, and some tokens are masked. The unmasked tokens are fed into the encoder, while the decoder predicts the masked tokens. During evaluation, these predicted tokens are decoded back to their original form using the decoder from the specific tokenizer for each modality.}
	\label{fig:method:mae-training}
\end{figure*}
Figure \ref{fig:method:mae-training} illustrates the overall architecture of the model. This figure shows that the tokenizers first decompose the input maps (modalities) into tokens. These tokens are then fed into the encoder and diffusion decoder \citep{diffusionDecoder} networks. The encoder network uses an input mask, while the diffusion decoder uses a target mask. The output embeddings produced by the encoder serve as context for the decoder in the cross-attention layers \citep{attention}, where the attention is defined as \citep{attention}:
\begin{equation}
	\mathrm{Attention}(Q, K, V) = \mathrm{softmax}\left(\frac{Q\cdot K^T}{\sqrt{d_k}}\right)\cdot V
\end{equation}
while the query matrix $Q$ is constructed using the decoder vectors, along with the key matrix $K$ and the value matrix $V$ from the encoder's output embeddings. This approach enables the diffusion decoder to predict the remaining target tokens by leveraging the context provided by the encoder.

The system is trained with a modality-wise loss function, which is defined as:
\begin{equation}
	\mathcal{L}_{\text{modality}} = \sum_{i=1}^{N_{\mathrm{mods}}} \mathrm{CE}(\hat{y}_{mod}\, ||\, y_{mod})
\end{equation}
where $\hat{y}_{mod}$ is the predicted output of the model, $y_{mod}$ is the target output, and $N_{\mathrm{mods}}$ is the number of modalities. The loss function is computed using the cross-entropy (CE) between the predicted and target tokens for each modality. This approach allows the model to learn from the relationships between different modalities and improve its performance in multi-modal tasks.

\section{Training and Evaluation}
\subsection{Training Pipeline}
One challenge in training a multimodal system is incorporating many modalities into one model. A promising approach is to pre-train the system using two modalities, such as grayscale images and DEMs, and then fine-tune it with additional modalities.

This pre-training on two modalities enables the system to learn their fundamental relationships. Once these relationships are established, the system can be fine-tuned with more modalities to enhance its performance.

We first pre-trained a smaller model using a sequence length of 1568 to train our model. This model randomly utilized either the grayscale tokens or the DEM (Digital Elevation Model) tokens as input, aiming to reproduce the corresponding modality. By doing so, we constructed an autoencoder that learns to integrate Shape from Shading and image rendering, allowing us to generate the DEM from the grayscale image or vice versa.

Once we have established this foundation, we incorporate the normal vectors into the model by adjusting the $\alpha$ values explained in Section \ref{sec:UnifiedTransformer}.
These values dictate the distribution of input modality tokens and are set to more meaningful levels based on the specific modality. Generally, we assign a low $\alpha$ value to the modalities Gray and DEM while the $\alpha$ value for the modality Normals is set higher. This approach allows the training to primarily include the Gray and DEM tokens while intermittently incorporating the Normals tokens into the model. As a result, the transformer continuously integrates knowledge about the normal vectors while maintaining the understanding of the grayscale to DEM transformation.

Lastly, we incorporate the Albedo feature into the transformer using the same strategy as outlined above. To enable the generation of one modality out of three given modalities, we increase the sequence length from 1568 to 2352 (calculated as 784 tokens per image multiplied by 3). We then copy the available weights from the smaller transformer to the larger one. After this, we maintain the previous configuration while adding a higher $\alpha$ value for the Albedo modality. This approach allows the Albedo modality to be incorporated into the transformer gradually, ensuring that the previous knowledge is preserved.

\subsection{Evaluation Metrics}
In this section, we will outline the metrics used to evaluate our multimodal any-to-any approach. To provide a complete assessment of our model's capabilities, the evaluation is divided into two complementary parts: qualitative and quantitative analysis. This dual approach is essential because while quantitative metrics offer objective and reproducible measurements of performance, qualitative evaluation provides indispensable insights into the model's behavior and the perceptual quality of its outputs. Combining these methods ensures that the limitations of one type of data are balanced by the strengths of another, leading to a more holistic understanding.

\subsubsection{Qualitative metrics}
To intuitively assess and understand the model's performance beyond numerical scores, we will first evaluate our results qualitatively. Quantitative metrics can sometimes fail to reflect perceptual quality, so this subjective assessment is important for judging the real-world applicability of the generated results. We will showcase various predictions to visually inspect the output quality, demonstrate cross-modal retrieval to assess the learned joint embedding space, and conduct ablation studies on specific data points to understand the influence of different components. Also, we will visualize the attention mechanism to provide further insights into the model's decision-making process.

\subsubsection{Quantitative metrics}
To objectively measure the performance of our model and enable thorough comparisons, we will conduct a quantitative evaluation. These metrics offer a standardized method for assessing the pixel-level accuracy of the generated images. The quantitative metrics are divided into two categories: pixel-wise and modal-wise. For the pixel-wise metrics, we will employ the following measures to compare the input image $x(i, j)$ with the predicted image $\hat{x}(i, j)$:
\begin{itemize}
	\item \textbf{Mean Squared Error (MSE)}: \begin{equation}\mathrm{MSE}(x, \hat{x}) = \frac{1}{mn}\sum_{i=1}^{n}\sum_{j=1}^{m}[x(i, j) - \hat{x}(i, j)]^2\end{equation}
	\item \textbf{Root Mean Squared Error (RMSE)}: \begin{equation}\mathrm{RMSE}(x, \hat{x}) = \sqrt{\mathrm{MSE}(x, \hat{x})}\end{equation}	
	\item \textbf{Peak Signal-to-Noise Ratio (PSNR)}: \begin{equation}\mathrm{PSNR}(x, \hat{x}) = 20\log_{10}\left( \frac{\mathrm{MAX}_x}{\sqrt{\mathrm{MSE}(x, \hat{x})}}\right)\end{equation}
	\item \textbf{Structural Similarity Index (SSIM)}: \begin{equation} \mathrm{SSIM}(x,y) = \frac{(2\mu_x\mu_y + c_1)(2\sigma_{xy} + c_2)}{(\mu_x^2 + \mu_y^2 + c_1)(\sigma_x^2 + \sigma_y^2 + c_2)}\end{equation}
	\item \textbf{Remaining Error (RE)}: (for DEM evaluation)\begin{equation}
		\mathrm{RE}_{<e\mathrm{m}}(x, \hat{x}) = \frac{1}{mn}\sum_{i=1}^{n}\sum_{j=1}^{m}\mathtt{int}(\left|x(i, j) - \hat{x}(i, j)\right|<e)
	\end{equation}
\end{itemize}
We will conduct two studies: an ablation study and a cross-modal retrieval study. These studies will help us evaluate the quality of the joint embedding space and assess the contributions of each modality to the model's performance. We will examine relevant metrics in various scenarios, including instances where one or more modalities are missing and a single remaining modality is used exclusively for prediction.

A special case in our evaluation involves DEMs. To measure the SSIM metric between two images, the values must be in the range of 0 to 1. Therefore, to compute the SSIM between the predicted and the ground truth DEM, we apply a Lambertian Reflectance model \citep{lambertShading} to shade both images and transform them into the 0 to 1 range, allowing us to simulate how the terrain would appear when illuminated from a specific direction. For this, we use an illumination vector of $[0.5; 0.0; 0.4]$.
\section{Results}
This section presents our findings regarding the multimodal transformer, detailing a progression from the experimental setup to thorough qualitative and quantitative analyses of the model's performance in reconstructing the lunar surface.

\subsection{Overview of Experimental Setup}
The main objective of these experiments is to assess the ability of a single, unified transformer architecture to learn shared representations and translate between various lunar data modalities. The model was designed to reconstruct missing data by utilizing the learned relationships among the available modalities. Consequently, two primary experimental scenarios were established: first, producing a single missing modality from the three available inputs, and second, generating three modalities from a single input modality.

\subsection{Qualitative Analysis}
This subsection presents a visual assessment of the model's performance, evaluating its ability to generate plausible lunar surface data under various conditions.
In the first scenario (see figure \ref{fig:results:mod_predictions_remaining}), where three modalities are provided to predict one missing modality, a visual analysis reveals a substantial reconstruction of the missing data. In this case, the model effectively produced every modality from the remaining modalities.

When tasked with generating three different types of data from a single input (see figure \ref{fig:results:mod_predictions_one2all}), the model produced varied but insightful results. For the first three input modalities, grayscale images, normal maps, and DEMs, the model achieved successful reconstruction. However, it struggled in the last case, where predictions were based on the albedo modality. This outcome was expected because albedo is primarily influenced by the surface composition and the effects of space weathering, making it independent of elevation and normal maps. This highlights the model's ability to separate albedo from geometric features.
\begin{figure*}
	\centering
	\begin{subfigure}[t]{0.95\textwidth}
		\centering
		\input{figures/tikz/results/mod_predictions.tex}
		\caption{Predicting one remaining modality from three input modalities}
		\label{fig:results:mod_predictions_remaining}
		\vspace*{0.75cm}
	\end{subfigure}
	\begin{subfigure}[t]{0.95\textwidth}
		\centering 
		\input{figures/tikz/results/mod_predictions_one2all.tex}
		\caption{Predicting three modalities from a single input modality}
		\label{fig:results:mod_predictions_one2all}
	\end{subfigure}
	\caption{Predictions of the multimodal model. The model generates in the first case one remaining modality while predicting in the second case all remaining modalities from a single one.}
	\label{fig:results:mod_predictions}
\end{figure*}

One effective way to explore the relationships between different modalities is by visualizing the attention mechanism. The Attention Map, which is created by computing the dot product of features, is generated from the network across all layers and then averaged over each pixel. Consequently, a single forward pass yields one Attention Map. To capture all relationships, we store the Attention Maps for all available test images and then average them across all pixels. Since the Transformer has an input sequence length of $2,352$ tokens (which equals $784\times 3$), we input all possible combinations of three inputs to predict the remaining one. This process results in the Attention Maps shown in Figure \ref{fig:results:attention_maps}.

The displayed attention maps highlight bright colors along the diagonal, while the rest of the matrix shows varying levels of brightness. This indicates that tokens within the same modality (represented by the diagonal boxes) are more highly correlated than those representing intermodality relationships. The off-diagonal elements illustrate how the model prioritizes information from the input modalities to make its predictions.

For example, when predicting the Gray modality (Figure \ref{fig:results:attention_maps_gray}), the model shows noteworthy attention to the Albedo and DEM inputs, indicating their importance for this task. In contrast, when predicting DEM (Figure \ref{fig:results:attention_maps_dem}), the model focuses primarily on the Normals data. The prediction for Normals (Figure \ref{fig:results:attention_maps_normals}) relies on both DEM and Gray inputs. Finally, in predicting Albedo (Figure \ref{fig:results:attention_maps_albedo}), the model gives the most attention to the Gray modality, which is reasonable since grayscale images are related to surface reflectance.

\begin{figure*}[!ht]
	\begin{subfigure}[t]{0.24\textwidth}
		\input{figures/png_results/attention_maps/attentionMap1.tex}
		\caption{Gray}
		\label{fig:results:attention_maps_gray}
	\end{subfigure}
	\begin{subfigure}[t]{0.24\textwidth}
		\input{figures/png_results/attention_maps/attentionMap2.tex}
		\caption{DEM}
		\label{fig:results:attention_maps_dem}
	\end{subfigure}
	\begin{subfigure}[t]{0.24\textwidth}
		\input{figures/png_results/attention_maps/attentionMap3.tex}
		\caption{Normals}
		\label{fig:results:attention_maps_normals}
	\end{subfigure}
	\begin{subfigure}[t]{0.24\textwidth}
		\input{figures/png_results/attention_maps/attentionMap4.tex}
		\caption{Albedo}
		\label{fig:results:attention_maps_albedo}
	\end{subfigure}
	\caption{The subfigures above present the summed attention maps for the available data points in the test dataset. The transformer model is evaluated using three different input modalities, aiming to predict the remaining modality. All attention maps generated from this process are stacked on top of one another and then summed, resulting in the four images displayed above.}
	\label{fig:results:attention_maps}
\end{figure*}

\subsection{Quantitative Analysis}
This subsection provides the numerical data and metrics used to evaluate our multimodal approach. The metrics are categorized into pixel-wise measures, which include Mean Squared Error (MSE), Root Mean Squared Error (RMSE), Peak Signal-to-Noise Ratio (PSNR), and Structural Similarity Index (SSIM).

We will first discuss the quantitative results for the scenario in which three modalities are used to predict a single remaining modality. Next, we will examine the results for the case where a single modality is employed to predict the other three modalities.

\subsubsection{Performance of Missing Modality Generation}
\begin{table}[!ht]
	\centering
	\begin{tabular}{l c c c c | c}
		\hline
		Modality: 						& MSE			& RMSE	 	& PSNR		& SSIM			& Datarange		\\\hline\hline
		\textcolor{MyBlue}{Gray} [I/F]	& $1.81\cdot10^{-06}$	& $0.00134$	& $37.2737$	& $0.726185$	& $0.09827$		\\
		\textcolor{MyGreen}{DEM} [m]	& $0.0756$		& $0.2751$	& $52.1714$	& $0.955542$	& $355.6104$	\\
		\textcolor{MyOrange}{Normals}	& $0.00023$	    & $0.0152$	& $41.4265$	& $0.917043$	& $2.0$			\\
		\textcolor{MyRed}{Albedo}		& $0.001195$	& $0.0345$	& $14.2597$	& $0.932345$	& $0.0801$		\\\hline
	\end{tabular}
	\caption{Quantitative pixelwise metrics for the multimodal model. For the case where three modalities are used to predict one remaining modality. The table shows the Mean Squared Error (MSE), Root Mean Squared Error (RMSE), Peak Signal-to-Noise Ratio (PSNR), and Structural Similarity Index (SSIM) for each predicted modality. Also shown is the data / value range for each modality of the test dataset.}
	\label{tab:results:metrics}
\end{table}
The results for the scenario predicting one modality from three are presented in Table \ref{tab:results:metrics}. The model shows varying levels of success, indicating that the relationships between the input modalities and the target modality differ in complexity and predictability.
Specifically, for the grayscale image, both the MSE and RMSE are low. However, it exhibits the lowest SSIM value, implying that while pixel-wise errors are minimal, there are slight discrepancies in the perceived structural details. This low SSIM may also be due to the absence of lighting information, such as the angle of light incidence, which is not currently included in the transformer model. As a result, shadows may be predicted differently, leading to structural inconsistencies despite the low MSE and RMSE.
The DEM has a much higher MSE/RMSE value, which is not surprising given its value range of several hundred meters. However, we observe the highest SSIM value, indicating that the model successfully captured the structural patterns and relative heights. For the Normal Vectors, the MSE and RMSE are in the moderate range, with a high SSIM. Lastly, for the Albedo map, the MSE and RMSE values were surprisingly high given its data range. However, the SSIM remained high, indicating that the algorithm effectively captured the spatial distribution and patterns of reflectivity.

\subsubsection{Performance of (from) Single Modality Generation}
\begin{figure*}[!ht]
	\begin{subfigure}[t]{0.45\textwidth}
		\centering
		\begin{tikzpicture}
    \begin{axis}[
        colormap={YlGn}{
            rgb255(0)=(255,255,229)
            rgb255(1)=(247,252,185)
            rgb255(2)=(217,240,163)
            rgb255(3)=(173,221,142)
            rgb255(4)=(120,198,121)
            rgb255(5)=(65,171,93)
            rgb255(6)=(35,132,67)
            rgb255(7)=(0,104,55)
            rgb255(8)=(0,69,41)
        },
        axis line style={draw=none},
        tick style={draw=none},
        xtick={0,1,2,3},
        xticklabels={gray $\leftarrow$, dem $\leftarrow$, normals $\leftarrow$, albedo $\leftarrow$},
        xticklabel style={rotate=35, anchor=east, yshift=-0.5em},
        ytick={0,1,2,3},
        yticklabels={gray $\rightarrow$, dem $\rightarrow$, normals $\rightarrow$, albedo $\rightarrow$},
        ymin=-0.5,
        ymax=3.5,
        xmin=-0.5,
        xmax=3.5,
        width=\textwidth,
        point meta=explicit,
        nodes near coords,
        nodes near coords align={center},
        nodes near coords style={font=\footnotesize},
        nodes near coords={%
            \pgfmathtruncatemacro{\rowy}{int(\coordindex/4)}%
            \pgfmathtruncatemacro{\colx}{int(mod(\coordindex,4))}%
            \ifnum\rowy=\colx
            \color{white}--
            \else
                \pgfmathparse{\pgfmathfloatvalueof{\pgfplotspointmeta} > 0.8 ? 1 : 0}%
                \ifnum\pgfmathresult=1
                    \color{white}\pgfmathprintnumber[fixed,precision=3]{\pgfplotspointmeta}
                \else
                    \color{black}\pgfmathprintnumber[fixed,precision=3]{\pgfplotspointmeta}
                \fi
            \fi
        }
    ]
        \addplot[matrix plot, mesh/rows=4, mesh/cols=4] table [x=x, y=y, meta=z] {
            x y z
            0 0 1.0 
            0 1 0.689
            0 2 0.758
            0 3 0.696
            1 0 0.7689267992973328
            1 1 1.0
            1 2 0.9770191311836243
            1 3 0.6622410416603088
            2 0 0.444
            2 1 0.953
            2 2 1.0 
            2 3 0.349
            3 0 0.952
            3 1 0.962
            3 2 0.955
            3 3 1.0
        };
    \end{axis}
\end{tikzpicture}
		\caption{SSIM (Structural Similarity Index)}
		\label{fig:heatmap:ssim}
	\end{subfigure}
	\hfill
	\begin{subfigure}[t]{0.45\textwidth}
		\centering
		\begin{tikzpicture}
    \begin{axis}[
        colormap={YlGn}{
             rgb255(0)=(255,255,229)
             rgb255(1)=(247,252,185)
             rgb255(2)=(217,240,163)
             rgb255(3)=(173,221,142)
             rgb255(4)=(120,198,121)
             rgb255(5)=(65,171,93)
             rgb255(6)=(35,132,67)
             rgb255(7)=(0,104,55)
             rgb255(8)=(0,69,41)
          },
        axis line style={draw=none},
        tick style={draw=none},
        xtick={0,1,2,3},
        xticklabels={gray $\leftarrow$, dem $\leftarrow$, normals $\leftarrow$, albedo $\leftarrow$},
        xticklabel style={rotate=35, anchor=east, yshift=-0.5em},
        ytick={0,1,2,3},
        yticklabels={gray $\rightarrow$, dem $\rightarrow$, normals $\rightarrow$, albedo $\rightarrow$},
        ymin=-0.5,
        ymax=3.5,
        xmin=-0.5,
        xmax=3.5,
        width=\textwidth,
        point meta=explicit,
        nodes near coords,
        nodes near coords align={center},
        nodes near coords style={font=\footnotesize},
        nodes near coords={%
            \pgfmathtruncatemacro{\rowy}{int(\coordindex/4)}%
            \pgfmathtruncatemacro{\colx}{int(mod(\coordindex,4))}%
            \ifnum\rowy=\colx
                \color{white}--
            \else
                \pgfmathparse{\pgfmathfloatvalueof{\pgfplotspointmeta} > 40 ? 1 : 0}%
                \ifnum\pgfmathresult=1
                    \color{white}\pgfmathprintnumber[fixed,precision=2]{\pgfplotspointmeta}
                \else
                    \color{black}\pgfmathprintnumber[fixed,precision=2]{\pgfplotspointmeta}
                \fi
            \fi
        }
    ]
        \addplot[matrix plot, mesh/rows=4, mesh/cols=4] table [x=x, y=y, meta=z] {
            x y z
            0 0 60 
            0 1 39.30171585083008
            0 2 39.95869064331055
            0 3 35.94383239746094
            1 0 27.424041748046875
            1 1 60.0 
            1 2 52.74962615966797
            1 3 24.05099105834961
            2 0 25.642370223999023
            2 1 44.00045394897461
            2 2 60.0 
            2 3 18.21569061279297
            3 0 13.115608215332031
            3 1 14.644335746765137
            3 2 12.53039264678955
            3 3 60.0
        };
    \end{axis}
\end{tikzpicture}
		\caption{PSNR (Peak Signal-to-Noise Ratio)}
		\label{fig:heatmap:psnr}
	\end{subfigure}
	\hfill
	\begin{subfigure}[t]{0.45\textwidth}
		\centering
		\begin{tikzpicture}
    \begin{axis}[
        colormap={YlGn}{
            rgb255(0)=(0,69,41)
            rgb255(1)=(0,104,55)
            rgb255(2)=(35,132,67)
            rgb255(3)=(65,171,93)
            rgb255(4)=(120,198,121)
            rgb255(5)=(173,221,142)
            rgb255(6)=(217,240,163)
            rgb255(7)=(247,252,185)
            rgb255(8)=(255,255,229)
        },
        axis line style={draw=none},
        tick style={draw=none},
        xtick={0,1,2,3},
        xticklabels={gray $\leftarrow$, dem $\leftarrow$, normals $\leftarrow$, albedo $\leftarrow$},
        xticklabel style={rotate=35, anchor=east, yshift=-0.5em},
        ytick={0,1,2,3},
        yticklabels={gray $\rightarrow$, dem $\rightarrow$, normals $\rightarrow$, albedo $\rightarrow$},
        ymin=-0.5,
        ymax=3.5,
        xmin=-0.5,
        xmax=3.5,
        width=\textwidth,
        point meta=explicit,
        nodes near coords,
        nodes near coords align={center},
        nodes near coords style={font=\footnotesize},
        nodes near coords={%
            \pgfmathtruncatemacro{\rowy}{int(\coordindex/4)}%
            \pgfmathtruncatemacro{\colx}{int(mod(\coordindex,4))}%
            \ifnum\rowy=\colx
            \color{white}--
            \else
                \pgfmathparse{\pgfmathfloatvalueof{\pgfplotspointmeta} < 4 ? 1 : 0}%
                \ifnum\pgfmathresult=1
                    \color{white}\pgfmathprintnumber[fixed,precision=4]{\pgfplotspointmeta}
                                    \else
                    \color{black}\pgfmathprintnumber[fixed,precision=3]{\pgfplotspointmeta}
                \fi
            \fi
        }
    ]
        \addplot[matrix plot, mesh/rows=4, mesh/cols=4] table [x=x, y=y, meta=z] {
            x y z
            0 0 0.0 
            0 1 0.001063311705365777
            0 2 0.04216029495000839
            0 3  0.0015615529846400023
            1 0 4.745567321777344
            1 1 0.0 
            1 2 0.25730472803115845
            1 3 6.989447593688965
            2 0 0.0936846062541008
            2 1 0.011343763209879398
            2 2 0.0 
            2 3 0.2201591432094574
            3 0 0.039412982761859894
            3 1 0.03316720947623253
            3 2 0.04216029495000839
            3 3 0.0
        };
    \end{axis}
\end{tikzpicture}
		\caption{RMSE (Root Mean Squared Error)}
		\label{fig:heatmap:rmse}
	\end{subfigure}
	\caption{Heatmap visualization of cross-modal relationships, when predicting \textit{all modalities from a single one}. All Heatmap show on the x-axis the predicted modality and on the y-axis the input modality. The first subfigure \ref{fig:heatmap:ssim} shows the SSIM metric, the second subfigure \ref{fig:heatmap:psnr} shows the PSNR metric, and the last subfigure \ref{fig:heatmap:rmse} shows the RMSE metric.}
	\label{fig:heatmap_grid}
\end{figure*}
This section evaluates the model's effectiveness in predicting remaining modalities using a single input modality. The metrics shown in Figure \ref{fig:heatmap_grid} show a substantial geometric correlation between the DEM and normals. Predictions from DEM to normals yield a high SSIM, high PSNR, and a low RMSE, whereas the reverse shows high SSIM and high PSNR but struggles with absolute heights, resulting in a higher RMSE. The albedo's weak correlation with geometry is evident, as predictions involving albedo show low SSIM and high RMSE values. The gray image acts as a complex composite signal incorporating geometry, reflectivity, and lighting, leading to successful predictions. However, predicting the DEM from grayscale presents challenges due to the ambiguity typical of Shape from Shading.
For the grayscale image, the RMSE is low across all inputs ($0.001063$ - $0.001561$). However, it has some of the lower SSIM values, suggesting that while pixel-wise errors are minimal, there are more noticeable inconsistencies in the perceived structural details compared to other outputs.
The DEM shows a much higher RMSE, which is expected given its value range spanning several hundred meters. Nonetheless, it boasts the highest single SSIM value ($0.977$ when predicted from normals), indicating that the model effectively captured the structural patterns and relative heights, particularly when geometric information from the normals is available.
When examining the normal vectors, we find that the RMSE is at a moderate level, while the model demonstrates a high SSIM. The SSIM is high ($0.953$) when predictions are made using the DEM, but it drops (to $0.444$ and $0.349$) when predictions are based on grayscale or albedo data. This indicates a strong dependence on the input geometry.
Finally, for the Albedo map, the RMSE values are relatively low and consistent ($0.039412$ - $0.042160$), but the SSIM remains elevated. In fact, predictions of albedo achieve the most consistently high SSIM scores across all inputs ($0.952$ - $0.962$), indicating that the algorithm successfully captured the spatial distribution and patterns of reflectivity regardless of the input modality.

\subsection{Shape and Albedo from Shading}
Our results demonstrate that multimodal learning can approach the Shape and Albedo from Shading problem by treating them as distinct modalities. The model's design learns the physical independence of photometric properties from topography. This is evident in the failure to generate geometry from albedo alone and is quantitatively confirmed by the weak prediction metrics between albedo and geometric modalities like DEMs and normals. Grayscale images serve as a complex composite signal combining the effects of geometry and photometric properties. The model's ability to separate albedo from 3D reconstruction validates a fundamental aspect of our unified model.

Directly comparing the accuracy of our DEM with that of other studies using the RMSE presents inherent challenges. As we have noted, RMSE is an absolute measure that is closely tied to the specific characteristics of our study site. Using these metrics at the highest available resolution often lacks a directly comparable ground truth dataset. Previous research \citep{tenthoff-2020,hess-2022} has tackled this issue by employing scale invariance, where Shape from Shading (SfS) is applied to lower-resolution imagery and validated against higher-resolution stereo data. As a result, a direct comparison using stereo data of the same high resolution or laser altimetry points may not provide the most meaningful evaluation of relative performance across different methodologies.
To compare the different prediction methods against the original ground truth (refer to the large area predictions in Figures \ref{fig:appendix:expected:dem} and \ref{fig:appendix:predicted:dem_from_gray} in the Appendix), we can examine the cross-sectional profiles.
When predicting the DEM using only the grayscale image, the profiles and gradients, illustrated in Figure \ref{fig:results:cross_sections:gray2dem}, show that the predictions (red line) align closely with the ground truth (dark blue line). However, noticeable mismatches occur at peaks and troughs, particularly around sharp changes in elevation. This is reflected in the metrics: $RE_{<2} = 0.33768$, $RE_{<4} = 0.63451$, and $RE_{<10} = 0.97733$.
In contrast, when using all three modalities (grayscale, normals, and albedo), the elevation and gradient profiles, shown in Figures \ref{fig:results:cross_sections:all2demElevation} and \ref{fig:results:cross_sections:all2demGradient}, respectively, effectively capture these sharp and abrupt changes with much higher accuracy. This substantial improvement is also evident in the $\mathrm{RE}_{<2m}$ metric, which reaches a value of $0.99867$.

\begin{table}
	\centering
	\begin{tabular}{l c c c c c}	
		\hline
		Method: 						 		& RMSE [m] 		& PSNR: 	& SSIM: 	& RE$_{<2}$: 	\\\hline\hline
		GADEM \cite{yang-2024-attention} 		& $2.498$ 	    & $52.3$ 	& $0.608$ 	& $0.616$ 		\\
		MadNet2 \cite{tao-2021-madnet2}  		& $1.0675$ 		& $-$ 	    & $0.894$ 	& $-$ 			\\
		Ours [gray$\rightarrow$dem] 	 		& $4.746$		& $27.42$ 	& $0.769$ 	& $0.33768$  	\\
		Ours [all$\rightarrow$dem]		 		& $0.2751$		& $52.1714$	& $0.956$	& $0.99867$		\\
		\hline
	\end{tabular}
	\caption{Quantitative comparison of DEM generation methods.}
	\label{tab:results:dem_comparison}
\end{table}
To provide a more standardized and comparable evaluation, we examined the SSIM. Our method achieved a high SSIM of up to $76.9\%$ (Grayscale to DEM prediction). This result indicates a substantial structural similarity when compared to other DEM generation approaches, which is higher than those of \cite{yang-2024-attention} with $60.8\%$ but lower than MadNet2 \citep{tao-2021-madnet2}, which achieved an SSIM value of $89.4\%$. The results are summarized in Table \ref{tab:results:dem_comparison}. In contrast to the other methods, which are highly specialised in the generation of DEMs from grayscale or RGB images, our method allows for more flexibility and shows that it can separate albedo information from geometry. However, this flexibility comes at the expense of lower performance in the DEM generation task, as indicated by the RMSE and PSNR values.

\begin{figure}
	\centering
	\begin{subfigure}[t]{0.95\textwidth}
		\centering
		\begin{tikzpicture}
    \begin{axis}[
        xlabel={distance [100 m]},
        ylabel={Elevation [m]},
        grid=both,
        minor grid style={gray!25},
        major grid style={gray!25},
        width=0.8\textwidth,
        height=0.3\textwidth,
        legend pos=south east,
        xmin=500, xmax=1500,
        xtick={500,600,700,800,900,1000,1100,1200,1300,1400,1500},
        xticklabels={2.5,3.0,3.5,4.0,4.5,5.0,5.5,6.0,6.5,7.0,7.5},
    ]
        \addplot[
            MyBlue!75!black,
            line width=1.25pt,
        ] table {figures/tikz/results/crossSections/data/all2demGtLineElevation.dat};
        
        \addplot[
            MyRed!75!white,
            opacity=0.7,
            line width=1.25pt,
        ] table {figures/tikz/results/crossSections/data/all2demPredLineElevation.dat};
        
        \legend{Ground Truth, Prediction}
    \end{axis}
\end{tikzpicture}
		\caption{Elevation of the DEM in case of predicting the DEM from all other modalities.}
		\label{fig:results:cross_sections:all2demElevation}
	\end{subfigure}
	\begin{subfigure}[t]{0.95\textwidth}
		\centering
		\begin{tikzpicture}
    \begin{axis}[
        xlabel={distance [100 m]},
        ylabel={Gradient [m/m]},
        grid=both,
        minor grid style={gray!25},
        major grid style={gray!25},
        width=0.8\textwidth,
        height=0.3\textwidth,
        legend pos=north east,
        xmin=500, xmax=1500,
        xtick={500,600,700,800,900,1000,1100,1200,1300,1400,1500},
        xticklabels={2.5,3.0,3.5,4.0,4.5,5.0,5.5,6.0,6.5,7.0,7.5},
    ]
        \addplot[
            MyBlue!75!black,
            line width=1.25pt,
        ] table {figures/tikz/results/crossSections/data/all2demGtLineGrad.dat};
        
        \addplot[
            MyRed!75!white,
            opacity=0.7,
            line width=1.25pt,
        ] table {figures/tikz/results/crossSections/data/all2demPredLineGrad.dat};
        
        \legend{Ground Truth, Prediction}
    \end{axis}
\end{tikzpicture}
		\caption{Gradient of the DEM in case of predicting the DEM from all other modalities.}
		\label{fig:results:cross_sections:all2demGradient}
	\end{subfigure}
	\begin{subfigure}[t]{0.95\textwidth}
		\centering
		\begin{tikzpicture}
    \begin{axis}[
        xlabel={distance [100 m]},
        ylabel={Elevation [m]},
        grid=both,
        minor grid style={gray!25},
        major grid style={gray!25},
        width=0.8\textwidth,
        height=0.3\textwidth,
        legend pos=south east,
        xmin=500, xmax=1500,
        xtick={500,600,700,800,900,1000,1100,1200,1300,1400,1500},
        xticklabels={2.5,3.0,3.5,4.0,4.5,5.0,5.5,6.0,6.5,7.0,7.5},
    ]
        \addplot[
            MyBlue!75!black,
            line width=1.25pt,
        ] table {figures/tikz/results/crossSections/data/gray2demGtLineElevation.dat};
        
        \addplot[
            MyRed!75!white,
            opacity=0.7,
            line width=1.25pt,
        ] table {figures/tikz/results/crossSections/data/gray2demPredLineElevation.dat};
        
        \legend{Ground Truth, Prediction}
    \end{axis}
\end{tikzpicture}
		\caption{Elevation of the DEM in case of predicting the DEM from grayscale images.}
		\label{fig:results:cross_sections:gray2demElevation}
	\end{subfigure}
	\begin{subfigure}[t]{0.95\textwidth}
		\centering
		\begin{tikzpicture}
    \begin{axis}[
        xlabel={distance [100 m]},
        ylabel={Gradient [m/m]},
        grid=both,
        minor grid style={gray!25},
        major grid style={gray!25},
        width=0.8\textwidth,
        height=0.3\textwidth,
        legend pos=north east,
        xmin=500, xmax=1500,
        xtick={500,600,700,800,900,1000,1100,1200,1300,1400,1500},
        xticklabels={2.5,3.0,3.5,4.0,4.5,5.0,5.5,6.0,6.5,7.0,7.5},
    ]
        \addplot[
            MyBlue!75!black,
            line width=1.25pt,
        ] table {figures/tikz/results/crossSections/data/gray2demGtLineGrad.dat};
        
        \addplot[
            MyRed!75!white,
            opacity=0.7,
            line width=1.25pt,
        ] table {figures/tikz/results/crossSections/data/gray2demPredLineGrad.dat};
        \legend{Ground Truth, Prediction}
    \end{axis}
\end{tikzpicture}
		\caption{Gradient of the DEM in case of predicting the DEM from grayscale images.}
		\label{fig:results:cross_sections:gray2demGradient}
	\end{subfigure}
	\caption{Cross-sections of the DEM predicted only from grayscale images. The first subfigures \ref{fig:results:cross_sections:all2demElevation} and \ref{fig:results:cross_sections:gray2demElevation} show the elevation profile, while the subfigures \ref{fig:results:cross_sections:all2demGradient} and \ref{fig:results:cross_sections:gray2demGradient} show the gradient profiles. The red line indicates the ground truth, and the black line indicates the prediction.}
	\label{fig:results:cross_sections:gray2dem}
\end{figure}

\section{Discussion}
This section interprets the results presented in the previous section, discusses their implications, and addresses the limitations of our work to outline future research directions.

\subsection{Interpretation of Findings}
Our results provide strong evidence for the effectiveness of a unified transformer architecture in learning shared representations between different lunar data modalities. The quantitative metrics demonstrate a strong geometric correlation between DEMs and surface normals, with predictions between these two modalities achieving high accuracy and structural similarity. This was expected, as normals are mathematically derived from the DEM's gradient, but the model's ability to learn this relationship so effectively validates its core learning mechanism.
A fundamental finding of the study is the model's ability to recognize that surface albedo is physically independent of topography. The model's qualitative inability to generate plausible geometric data based solely on albedo, combined with weak quantitative metrics, indicates that the model can learn valid correlations. Instead, the model comprehended that albedo is related to material composition rather than shape. This outcome confirms that our multimodal approach effectively addresses the classic, ill-posed problem of inferring shape and albedo from shading by treating shape and albedo as separate outputs.

\subsection{DEM Generation from a Single Grayscale Image}
An important test of our model was the generation of a DEM from a single grayscale image, a task central to planetary science. Our model achieved a high SSIM, indicating that it successfully preserved the texture and relative structures of the lunar surface. However, the high RMSE and the visible deviations in elevation profiles, particularly at peaks and troughs, highlight a persistent challenge: accurately predicting absolute heights and the overall low-frequency topographic component from shading information alone.
It is important to place this result in the appropriate background. Many contemporary machine learning methods for generating DEMs explicitly use a low-resolution DEM as an additional input alongside a grayscale image. This initial DEM provides basic low-frequency information, allowing the model to concentrate on inferring high-frequency details from the image. 
In our study, the low-resolution DEM was utilized exclusively during the data preparation and generation stages. It served two primary purposes: first, to aid in the stitching of large-area image mosaics, and second, to provide a low-frequency constraint for the Shape from Shading method that created our ground truth data. However, the transformer model, during the process of predicting the DEM from the grayscale image, was never given this low-frequency element as input. Instead, it faced the much more challenging task of reconstructing the entire DEM, including both its low-frequency shape and high-frequency details, using solely a single grayscale image. Furthermore, unlike the SfS method, the model did not have access to the illumination geometry and had to infer the relationship from the data itself.
The fact that the model was able to reconstruct the structure so accurately (as indicated by a high SSIM) suggests that it has learned the fundamental, physically-based relationships between shading and shape.

\subsection{Contribution and Implications}
This work successfully demonstrates the potential of multimodal learning to unify disparate computer vision tasks in lunar remote sensing within a single, flexible model. Rather than training specialized models for each task, our any-to-any model can perform 3D reconstruction, albedo estimation, and data completion by using learned cross-modal relationships. This holistic approach is particularly valuable for planetary science, where datasets are often incomplete, derived from different instruments, or have sparse spatial coverage. Our findings validate multimodal learning as a powerful approach for analyzing complex planetary data, mainly when working with incomplete datasets.

\subsection{Limitations and Future Work}
The primary limitation of our current model is the difficulty in accurately predicting absolute DEM heights from a single grayscale image without an explicit low-frequency prior, as discussed above. While our model captures relative topography well, the lack of an absolute height reference during inference leads to larger errors compared to methods that use an initial low-resolution DEM.
Future work will focus on expanding this unified model. An immediate step is to incorporate additional modalities, such as illumination and viewing geometry, as inputs. Providing the model with explicit information about the Sun's position could improve its ability to interpret shadows and shading, thereby enhancing DEM accuracy from grayscale images. Finally, this model can be extended to tackle more complex tasks such as photometric normalization and the co-registration of data from different missions to increase its power and flexibility.

\section{Conclusion}
This work demonstrates the potential of multimodal learning to unify computer vision tasks in lunar remote sensing. Our model successfully reconstructs missing modalities by exploiting cross-modal relationships between grayscale images, DEMs, surface normals, and albedo maps, effectively performing tasks like 3D reconstruction and photometric parameter estimation. Our qualitative results demonstrate strong reconstruction capabilities, which are particularly evident when predicting from three modalities to one remaining modality. Further, our quantitative metrics (SSIM, PSNR, and RMSE) demonstrate a strong geometric correlation between DEMs and normals, with predictions between these two modalities achieving high accuracy. This strong dependence was consistent across all metrics. In contrast, albedo exhibits a limited correlation with geometric modalities, highlighting the model's ability to separate albedo from 3D reconstruction. 
Furthermore, it became clear that grayscale images serve as a complex composite signal combining effects of illumination geometry and inherent reflectance. While high SSIM values generally confirm the preservation of crucial structural details, we found that our model can effectively generate a DEM from a single grayscale image, validating a key aspect of our unified framework. However, challenges persist in accurately predicting absolute DEM heights and albedo maps. These findings validate the multimodal learning approach for analyzing lunar data, particularly when working with incomplete datasets. Future work will expand this unified framework by incorporating additional modalities to tackle tasks such as photometric normalization and co-registration.

\section*{Declaration of Generative AI and AI-assisted technologies in the writing process}
During the preparation of this work the author(s) used Grammarly AI in order to improve the clarity and readability of the text. After using this tool/service, the author(s) reviewed and edited the content as needed and take(s) full responsibility for the content of the publication.



\appendix
\section*{Appendix}
\label{sec:appendix}
This appendix provides additional details on the data handling, tokenizer architecture, and model architecture used in this study. The sections are organized as follows:
\begin{itemize}
    \item \ref{sec:appendix_data}: Data handling and preparation
    \item \ref{sec:appendix_tokenizer}: Tokenizer architecture and training details
    \item \ref{sec:appendix_model}: Model architecture and training details
    \item \ref{appendix:SfS}: Shape from Shading regularization terms
\end{itemize}

\section{Data handling and preparation}
\label{sec:appendix_data}
Normalization is a crucial step in preparing the data for training and evaluation. The following normalization techniques were applied to different data types:
\begin{itemize}
    \item \textbf{Gray}: \texttt{torchvision $\rightarrow$ Normalize(...)}
    \item \textbf{DEM}: Subtracting mean from each image
    \item \textbf{Normals}: \texttt{None}
    \item \textbf{Albedo}: \texttt{torchvision $\rightarrow$ Normalize(...)}
\end{itemize}
All data types were stroed in pixelwise ontop of each other.

\section{Tokenizer Architecture and Training details}
\label{sec:appendix_tokenizer}
\begin{table}
    \centering
    \begin{tabular}{l | c c c c}
        \hline
        \textbf{Parameter:}     & \multicolumn{4}{c}{\textbf{Modality}} \\
                                & \textcolor{MyBlue}{\textbf{Gray}} & \textcolor{MyGreen}{\textbf{DEM}}  & \textcolor{MyOrange}{\textbf{Normals}}  & \textcolor{MyRed}{\textbf{Albedo}} \\
        \hline\hline
        vocab\_size             &  $1536$        &  $2048$       & $1536$            & $1024$ \\
        decay                   & \multicolumn{4}{c}{$0.99$}    \\
        use\_ema                & \multicolumn{4}{c}{\texttt{True}}   \\        
        \hline
        imgSize                 & \multicolumn{4}{c}{$224\times 224$}  \\
        patch\_size             & \multicolumn{4}{c}{$8\times 8$} \\   
        \hline
        embedding\_dim          & \multicolumn{4}{c}{$128$}   \\
        enc\_num\_heads         & \multicolumn{4}{c}{$4$}     \\
        enc\_num\_layers        & \multicolumn{4}{c}{$6$}     \\
        \hline
        dec\_conv\_dim          & \multicolumn{4}{c}{$64$}      \\
        \hline
        \#parameters            & $1.5$M        &  $1.6$M       & $1.5$M            & $1.5$M  \\
        \hline
    \end{tabular}
    \caption{Parameters of tokenizer architecture for all modalities. The parameters are the same for all modalities, except for the vocab\_size.}
    \label{tab:appendix:tokenizerParameters}
\end{table}
All tokenizer architectures share a common structure that incorporates a Vision Transformer (ViT) encoder, which is configured identically across all modalities. This includes parameters such as embedding dimension, number of attention heads, and number of transformer layers (see Table \ref{tab:appendix:tokenizerParameters}). A consistent upsampling convolutional neural network (UpConv CNN) serves as the decoder for each modality (see Figure \ref{fig:appendix:UpConvBlock} for each UConv Block structure). Additionally, the central Vector Quantization (VQ) layer maintains the same decay rate and uses an exponential moving average (EMA). 

\begin{figure}
    \centering
    \begin{tikzpicture}
    \draw[rounded corners, thick, fill=MyBlue!15!white, draw=MyBlue!15!black] (-2, -1.5) rectangle (1.25, 1.5);

    \draw[->, thick] (-2.25, 0) -- (-1.75, 0);
    \draw[->, thick] (-1.25, 0) -- (-1.0, 0);
    \draw[->, thick] (-0.5, 0) -- (-0.25, 0);
    \draw[->, thick] (0.25, 0) -- (0.5, 0);
    \draw[->, thick] (1.0, 0) -- (1.5, 0);

    \draw[rounded corners, fill=MyBlue!50!white] (-1.75, -1.3) rectangle (-1.25, 1.3) node[anchor=center, pos=.5, rotate=90] {\footnotesize \texttt{ConvTranspose2D}};
    \draw[rounded corners, fill=MyBlue!50!white] (-1.0, -1.3) rectangle (-0.5, 1.3) node[anchor=center, pos=.5, rotate=90] {\footnotesize \texttt{Conv2D}};
    \draw[rounded corners, fill=MyBlue!50!white] (-0.25, -1.3) rectangle (0.25, 1.3) node[anchor=center, pos=.5, rotate=90] {\footnotesize \texttt{BatchNorm2d}};
    \draw[rounded corners, fill=MyBlue!50!white] (0.5, -1.3) rectangle (1.0, 1.3) node[anchor=center, pos=.5, rotate=90] {\footnotesize \texttt{LeakyReLU}};
\end{tikzpicture}
    \caption{This figure shows a single decoder building block consisting of a ConvTranspose2D layer, followed by a Conv2D layer, BatchNormalization2D, and a LeakyReLU activation function.}
    \label{fig:appendix:UpConvBlock}
\end{figure}

The main difference between the tokenizers is the size of their vocabularies, which are customized to suit the unique characteristics and complexity of each input modality: grayscale images, Digital Elevation Models (DEMs), surface normals, and albedo maps. This tailored approach to vocabulary size enables each tokenizer to effectively learn a discrete latent space that is both efficient and representative of its specific data type while still utilizing the shared encoder-decoder backbone for consistent feature extraction and reconstruction.

Furthermore, all tokenizers are trained with a batch size of $128$ and $64$ epochs, as shown in Table \ref{tab:appendix:tokenizerTrainingParameters}.
\begin{table}
    \centering
    \begin{tabular}{c c}
        \hline
        \textbf{Parameter:} & \textbf{Value:} \\
        \hline\hline
        \texttt{batch\_size} & $128$ \\
        \texttt{num\_epochs} & $64$ \\
        \hline
    \end{tabular}
    \caption{Training parameters of the tokenizer.}
    \label{tab:appendix:tokenizerTrainingParameters}
\end{table}

\section{Model Architecture and Training details}
\label{sec:appendix_model}
Table \ref{tab:appendix:modelParameters} presents a detailed overview of the key hyperparameters used in training the multimodal model that follows the 4M Architecture. The optimization process utilized the AdamW optimizer \citep{adamW} with $\beta$ parameters set to $(0.9, 0.99)$. The initial learning rate was set to $5 \cdot 10^{-5}$, and a weight decay of $0.05$ was applied. To ensure training stability, gradient clipping was implemented with a maximum norm of $3.0$. The learning rate schedule adhered to a Cosine Annealing strategy \citep{cosineLR}, featuring a maximum cycle length (Tmax) of 5 epochs and a minimum learning rate of $5 \cdot 10^{-8}$.

The model's embedding dimension was established at 384, with embedding weights initialized using a normal distribution that has a standard deviation of 0.15. The loss function used was of the 'mod' type (the specific formulation of this loss is elaborated in Section \ref{sec:method:maskedAutoencoder}). Both the number of input and target tokens was set to 2358.

Within the Transformer architecture of the 4M model, 6 attention heads were utilized in the multi-head attention layers. The encoder and decoder components each consisted of 6 layers, with bias terms explicitly disabled in the projection (proj\_bias) and query-key-value (qkv\_bias) layers. Additionally, a gated multi-layer perceptron (gated\_mlp) was incorporated within the feed-forward network of the Transformer blocks. The model made use of pre-trained weights according to the 4M architecture\footnote{We modified the code of the 4M Model for our experiments. The code of the original model can be found on GitHub: \url{https://github.com/apple/ml-4m/}} (\texttt{pretrained4M = True}).

In total, the resulting model contains $29.7$ million trainable parameters, with $10.6$ million in the encoder, $14.2$ million in the decoder, and $2.4$ million in each of the encoder and decoder embedding layers.
\begin{table}
    \centering
    \begin{tabular}{l l}
        \hline
        \textbf{Parameter:}     & \textbf{Value:} \\
        \hline\hline
        optimizer               & AdamW \\
        betas                   & $(0.9, 0.99)$ \\
        learning\_rate          & $5\cdot10^{-5}$ \\
        weight\_decay           & $0.05$ \\
        gradient\_clip\_value   & $3.0$ \\
        scheduler               & \texttt{CosineAnnealingLR(...)} \\
        \,\,\rotatebox[origin=c]{180}{$\Lsh$} T\_max   & $5$ \\
        \,\,\rotatebox[origin=c]{180}{$\Lsh$} eta\_min            & $5\cdot10^{-8}$ \\
        \hline
        embed\_dim              & $384$ \\
        embedding\_init\_std    & $0.15$ \\
        \hline
        loss\_type              & mod \\
        num\_input\_tokens      & $2352$ \\
        num\_target\_tokens     & $2352$ \\
        \hline
        num\_heads              & $6$ \\
        encoder\_depth          & $6$ \\
        decoder\_depth          & $6$ \\
        proj\_bias              & \texttt{False} \\
        qkv\_bias               & \texttt{False} \\
        gated\_mlp              & \texttt{True} \\
        pretrained4M            & \texttt{True} \\
        \hline
        \#encoder\_parameters   & $10.6$M \\
        \#decoder\_parameters   & $14.2$M \\
        \#encoder\_embed\_parameters   & $2.4$M \\
        \#decoder\_embed\_parameters   & $2.4$M \\
        \#total\_parameters     & $29.7$M \\
        \hline
    \end{tabular}
    \caption{Hyperparameters of the multimodal model.}
    \label{tab:appendix:modelParameters}
\end{table}

\section{Shape from Shading Regularization Terms}
\label{appendix:SfS}
Because the choice of regularization terms affects the results of the SfS algorithm, we provide a brief description of the regularization terms used in this work.
We adopt the integrability error 
\begin{equation}
    E_\textnormal{int} = \frac{1}{2} \int\limits_{x}\int\limits_{y} \left(\frac{\partial z}{\partial x} - p\right)^2 + \left(\frac{\partial z}{\partial y} - q\right)^2 dx dy 
\end{equation}
introduced by \cite{horn1990-sfs} to enforce integrability of the noisy gradient field.

Shape from Shading can accurately estimate the surface slope, but the absolute height is not well constrained without a suitable initialization.
We restrict the deviation of the estimated DEM from the initial DEM $z_\textnormal{DEM}$ \citep{barker-2016} with two additional regularization terms introduced by \citep{dem-grumpe2014}. 
The relative depth constraint is given by
\begin{equation}
    \begin{split}
    E_\textnormal{rel} = &\frac{1}{2} \int\limits_{x}\int\limits_{y} \left(f(p) - f\left(\frac{\partial z_\textnormal{DEM}}{\partial x}\right)\right)^2 \\
    &+ \left(f(q) - f\left(\frac{\partial z_\textnormal{DEM}}{\partial y}\right)\right)^2 dx dy
\end{split}
\end{equation}
with a low-pass filter $f$ to match the image resolution to the lower DEM resolution.

The absolute depth constraint 
\begin{equation}
    E_\textnormal{abs} = \frac{1}{2} \int\limits_{x}\int\limits_{y} \left(f(z) - f(z_\textnormal{DEM})\right)^2 dx dy
\end{equation}
ensures that the low-frequency components of the DEM are similar to the initial DEM.
The total error to be minimized is the weighted sum of the individual error terms:
\begin{equation}
    E_\textnormal{total} = E_\textnormal{I} + \gamma E_\textnormal{int} + \delta E_\textnormal{rel} + \tau E_\textnormal{abs}
\end{equation}

\section{Larger DEM Areas}
\label{appendix:LargeDEM}

\begin{figure}
    \centering
    \input{figures/tikz/results/largeArea/inputDEM.tex}
    \caption{Expected DEM (ground truth). The red dash-dotted line indicates where the profile was extracted for comparison}
    \label{fig:appendix:expected:dem}
\end{figure}

\clearpage
\begin{figure}
    \centering
    \input{figures/tikz/results/largeArea/predDEMfromAll.tex}
    \caption{Predicted DEM from all other modalities (gray, normals, albedo). The red dash-dotted line indicates where the profile was extracted for comparison}
    \label{fig:appendix:predicted:dem_from_all}
\end{figure}

\clearpage
\begin{figure}
    \centering
    \input{figures/tikz/results/largeArea/predDEMfromGray.tex}
    \caption{Predicted DEM from Grayscale Image. The red dash-dotted line indicates where the profile was extracted for comparison}
    \label{fig:appendix:predicted:dem_from_gray}
\end{figure}

\bibliographystyle{elsarticle-harv}
\bibliography{bibliography/journal}

\vfill

\end{document}